\documentclass[lettersize,journal]{IEEEtran}
\usepackage{amsmath,amsfonts}
\usepackage{array}
\usepackage[caption=false,font=normalsize,labelfont=sf,textfont=sf]{subfig}
\usepackage{booktabs}
\usepackage{textcomp}
\usepackage{multirow} 
\usepackage{array} 
\usepackage{wrapfig}
\usepackage{stfloats}
\usepackage{url}
\usepackage{verbatim}
\usepackage{graphicx}
\usepackage{cite}
 \usepackage{algorithm}
 \usepackage{algorithmicx}
 \usepackage{color} 

\usepackage{hyperref}
 \usepackage{algpseudocode}
\begin{document}

\title{J-DDL: Surface Damage Detection and Localization System for Fighter Aircraft}
% \title{S3DLAS: A Smart System for Damage Detection and 3D Localization on Large Aircraft Surface}

\author{
 Jin Huang, Mingqiang Wei,~\IEEEmembership{Senior Member,~IEEE}, Zikuan Li, Hangyu Qu, Wei Zhao, and Xinyu Bai
        % <-this % stops a space
\thanks{Jin Huang and Mingqiang Wei are with School of Computer Science and Technology, Nanjing University of Aeronautics and Astronautics, Nanjing, China, and also with School of Artificial Intelligence, Taiyuan University of Technology, Taiyan, China (e-mail: jinhuang.nuaa@gmail.com, mingqiang.wei@gmail.com).} 
\thanks{Zikuan Li and Wei Zhao are with  School of Computer Science and Technology, Nanjing University of Aeronautics and Astronautics, Nanjing, China (e-mail: lizikuanhhu@gmail.com,   weizhao0120@nuaa.edu.cn).}
% <-this % stops a space
\thanks{Hangyu Qu is with Nanjing Institute of Technology, China (e-mail: y00450231019@njit.edu.cn).}
\thanks{Xinyu Bai is with Avic Shenyang Aircraft Company Limited, Shenyang, China (e-mail: nationlismbai@163.com).}
}

\markboth{Journal of \LaTeX\ Class Files,~Vol.~14, No.~8, August~2021}%
{Shell \MakeLowercase{\textit{et al.}}: A Sample Article Using IEEEtran.cls for IEEE Journals}


\maketitle

\begin{abstract}
% Ensuring the safety and extended operational life of fighter aircraft requires regular and thorough inspections. However, accurately detecting and localizing damage on aircraft surfaces remains a significant challenge due to the scale, complexity, and structural variations of their exteriors.
Ensuring the safety and extended operational life of fighter aircraft necessitates frequent and exhaustive inspections. While surface defect detection is feasible for human inspectors, manual methods face critical limitations in scalability, efficiency, and consistency due to the vast surface area, structural complexity, and operational demands of aircraft maintenance.
We propose a smart surface damage detection and localization system for fighter aircraft, termed J-DDL.  J-DDL integrates 2D images and 3D point clouds of the entire aircraft surface, captured using a combined system of laser scanners and cameras, to achieve precise damage detection and localization.
Central to our system is a novel damage detection network built on the YOLO architecture, specifically optimized for identifying surface defects in 2D aircraft images. Key innovations include lightweight Fasternet blocks for efficient feature extraction, an optimized neck architecture incorporating Efficient Multiscale Attention (EMA) modules for superior feature aggregation, and the introduction of a novel loss function, Inner-CIOU, to enhance detection accuracy.
After detecting damage in 2D images, the system maps the identified anomalies onto corresponding 3D point clouds, enabling accurate 3D localization of defects across the aircraft surface. Our J-DDL not only streamlines the inspection process but also ensures more comprehensive and detailed coverage of large and complex aircraft exteriors. To facilitate further advancements in this domain, we have developed the first publicly available dataset specifically focused on aircraft damage. Experimental evaluations validate the effectiveness of our framework, underscoring its potential to significantly advance automated aircraft inspection technologies. 
\end{abstract}

% \def\abstractname{Note to Practitioners}
% \begin{abstract}
% This paper presents a novel system, J-DDL, for detecting and localizing surface damage on fighter aircraft. The system leverages advanced computer vision and deep learning techniques to automate the inspection process, reducing the need for manual checks and improving accuracy.  This work is particularly relevant for aerospace engineers and maintenance crews seeking to enhance the efficiency and reliability of aircraft inspection processes.
% \end{abstract}


\section{Introduction}
\label{introduction}
\IEEEPARstart{D}{a}{m}{a}{g}{e} detection is a fundamental aspect of fighter aircraft maintenance, governed by strict standards to ensure safety, accuracy, and operational efficiency. 
Damage in the form of cracks, scratches, or others can severely compromise the structural integrity of an aircraft, leading to potential safety hazards. 
Currently, inspection methods are predominantly manual, relying on the expertise of human operators to visually assess aircraft surfaces~\cite{yasuda2022aircraft}. 
The most commonly employed procedure is General Visual Inspection (GVI), playing a crucial role in the aviation industry by providing a baseline assessment of aircraft performance.
However, conducting thorough inspections on large aircraft poses significant challenges due to the vast surface area that must be monitored. 
The reliance on manual techniques, such as direct visual assessments or the use of mirrors, magnifying lenses, and endoscopes~\cite{papa2018preliminary}, often proves inadequate for large-scale inspections, as these methods are time-consuming, labor-intensive, and prone to human error~\cite{sheikhalishahi2016human}. 
The inherent subjectivity of human-based inspections further exacerbates the difficulty in consistently identifying small defects across extensive airplane surfaces.

With the rapid development of sensing techniques and machine learning processing~\cite{wei2017_tase,chen2021_tase,HuangCity3d_2022},  some automated damage detection systems are proposed to address the limitations of manual inspection. 
These systems typically utilize unmanned aerial vehicles (UAVs) equipped with high-resolution cameras to capture detailed 2D images of extensive aircraft surfaces~\cite{liu2022uav, shao2023aircraft,zhang2024semi}. 
After image acquisition, learning-based algorithms are employed to identify anomalous regions, while localization modules precisely pinpoint the defects on the aircraft surfaces.
Despite these advancements, achieving fully automated and highly efficient damage detection for large aircraft surfaces remains a significant challenge, with several key issues yet to be resolved:
\begin{itemize}
\item How can the detection process be conducted without risking damage to the aircraft, particularly considering the possibility of UAV malfunctions that may lead to collisions with the airplane surface?
\item How to effectively detect damage across the vast and complex surfaces of large airplanes, considering the limited availability of comprehensive defect data to train detection models?
\end{itemize}

\begin{figure*}
    \centering
    \includegraphics[width=0.95\linewidth]{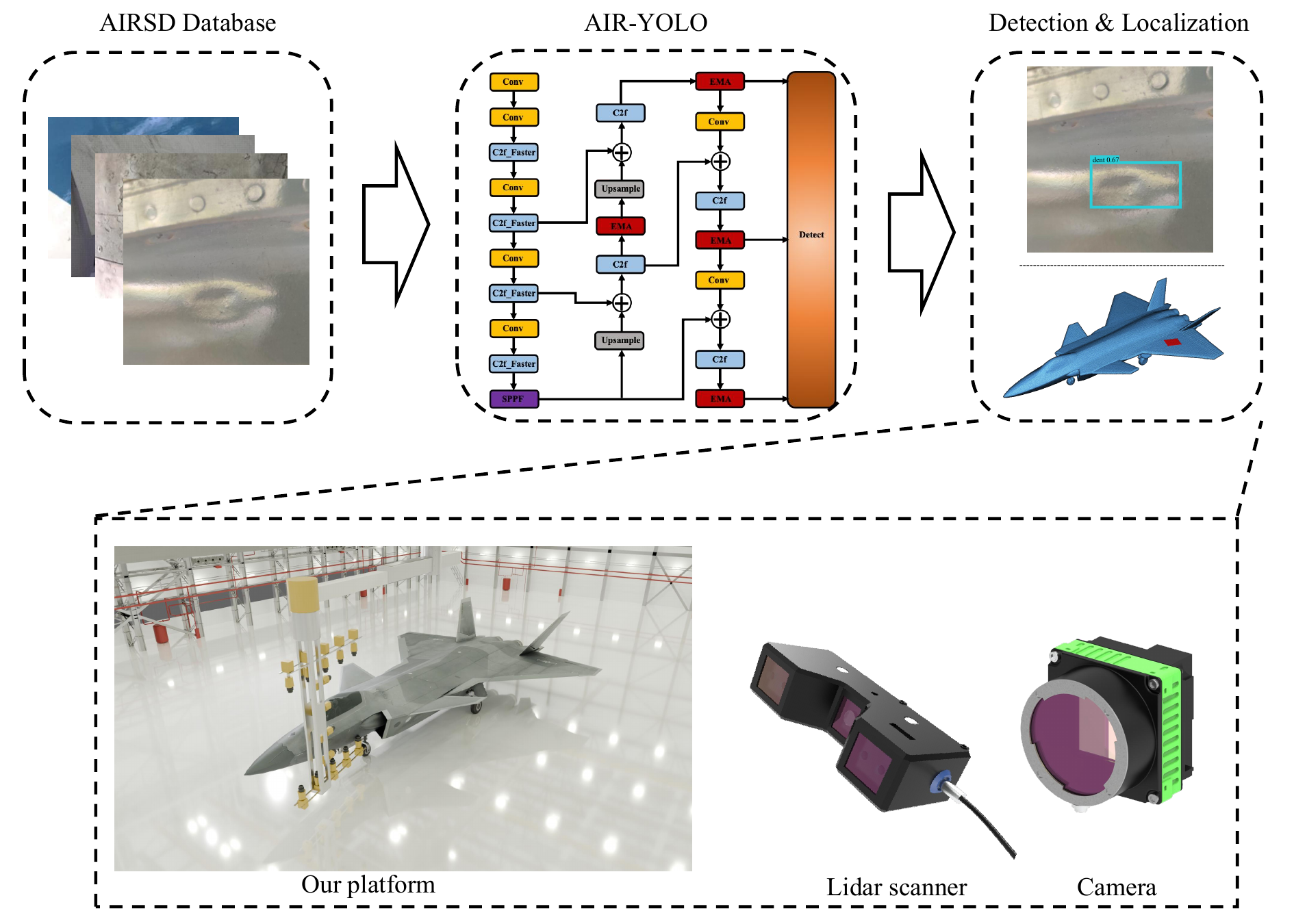}
    \caption{Overview of our proposed framework. The designed platform captures 2D images and 3D point clouds of the airplane. Next, the AIR-YOLO trained from AIRSD database is employed to accurately detect and classify anomalous regions. Lastly, the identified damage areas are localized within the 3D point cloud, providing critical spatial information to assist the human inspector.}
    \label{fig:pipeline}
\end{figure*}

These challenges underscore the need for methods capable of detecting and localizing damage across the entirety of an aircraft’s surface while maintaining operational efficiency.
Therefore, we propose a novel framework to detect and localize damage on large airplane surfaces, utilizing a specially designed device, as illustrated in Fig.\ref{fig:pipeline}. 
To avoid the risk of UAVs potentially colliding with and damaging aircraft surfaces, our system incorporates mechanically stabilized supports for mounting cameras and laser scanners. 
This design ensures precise and secure operation during inspections. 
The device is engineered to traverse the aircraft's silhouette each time the plane taxis into the hangar, systematically capturing high-resolution 2D and 3D representations of the surface. 
To achieve comprehensive coverage, the system integrates cameras and laser scanners positioned at multiple heights and angles, enabling the effective capture of both the upper and lower regions of the aircraft. 
Unlike traditional methods, which often neglect these critical areas, our configuration ensures thorough inspection from diverse perspectives and altitudes, thereby enhancing the accuracy and completeness of the imaging process
Furthermore, we introduce a novel detection network based on an improved YOLOv8~\cite{Jocher_Ultralytics_YOLO_2023} architecture to enhance the accuracy of damage detection. 
Finally,  the detected damage is localized within the 3D point clouds generated by the system. 
To support and advance the development of airplane defect detection, we also construct a large image dataset (AIRSD) specifically for airplane surface damage.
Our  contributions are mainly three-fold:
\begin{itemize}
    \item We propose a novel framework for automated detection and 3D localization of airplane surface damage, utilizing a combined laser scanning and photographic platform. 
    This system provides high-accuracy damage detection across the entire airplane surface, significantly improving efficiency compared to traditional methods.
    \item We present AIR-YOLO, a detection network specifically designed for airplane surface damage detection. 
    The network features a lightweight backbone, replacing standard convolutional layers with Fasternet blocks, which significantly reduces the number of model parameters. Additionally, the network incorporates an optimized neck architecture with Efficient Multi-Scale Attention (EMA) modules to enhance feature aggregation and introduces an Inner-IoU loss function to improve detection accuracy.
    \item We construct AIRSD\footnote{\url{https://drive.google.com/drive/folders/1BdbUv-vwLiNKtsd78rg5QYlSLqnzBOLX?usp=drive_link}}, the first dataset for fighter aircraft surface damage. 
    It comprises both synthetic and real-world data to ensure more comprehensive coverage of various damage scenarios. 
    AIRSD contains 8,091 images, each with a resolution of  $640 \times 640$, and covers 11 types of airplane surface damage.
    It serves as a valuable resource for advancing research and development in the field of automated aircraft inspection, providing a robust foundation for training and evaluating damage detection models.
\end{itemize}

\section{Related work}
\label{related work}
Since there is a large number of methods related to the field of airplane maintenance and inspection.
In this section, we mainly focus on the fields of fuselage inspection, object detection, and damage detection.

\subsection{Fuselage Inspection}
In the past, aircraft surface inspections have relied heavily on human visual assessment. 
Several studies have evaluated the reliability of human inspectors in identifying structural defects in aircraft~\cite{rummel1989applications,spencer1995reliability,murgatroyd1995study}. 
Aircraft maintenance and inspection tasks are often conducted within complex operational environments, characterized by time constraints, limited feedback, and challenging ambient conditions. 
These factors, coupled with inherent human error tendencies, contribute to a range of potential inspection errors~\cite{latorella2017review}.

To enhance both the efficiency and accuracy of fuselage inspections, various automated robotic systems have been proposed to localize damage more effectively. 
For example, White et al.~\cite{white2005mobile} introduce a novel clinging robot designed for aircraft surface inspection, which employs a specialized traction system to enable rapid movement across both planar and curved surfaces. 
Shang et al.~\cite{shang2007design}  develop a climbing robot designed for non-destructive testing (NDT) on aircraft surfaces. 
The robot employs vacuum suction cups for adhesion, pneumatic cylinders for moving, and a central rotation mechanism for directional corrections, with a flexible yet rigid structure to adapt to varying surface curvatures and ensure reliable inspections.
Colin et al.\cite{colin2018collaborative} introduce Air-Cobot, which employs a ground robot for aircraft inspection. 
The robot is equipped with a camera and a 3D laser scanner, enabling it to capture 2D images and 3D spatial data, as demonstrated in prior works\cite{futterlieb2014navigational, lakrouf2017moving}.

The necessity of contact with the fuselage surface increases the risk of causing damage to the aircraft, and the robot’s limited mobility prevents it from inspecting the top surfaces of the aircraft.
To address these limitations, we propose a novel hardware platform that offers improved control and versatility, enabling thorough inspection of both the upper and lower surfaces of the aircraft, without the risk of surface damage.

\begin{figure*}[htbp]
    \centering
    \includegraphics[width=0.72\linewidth]{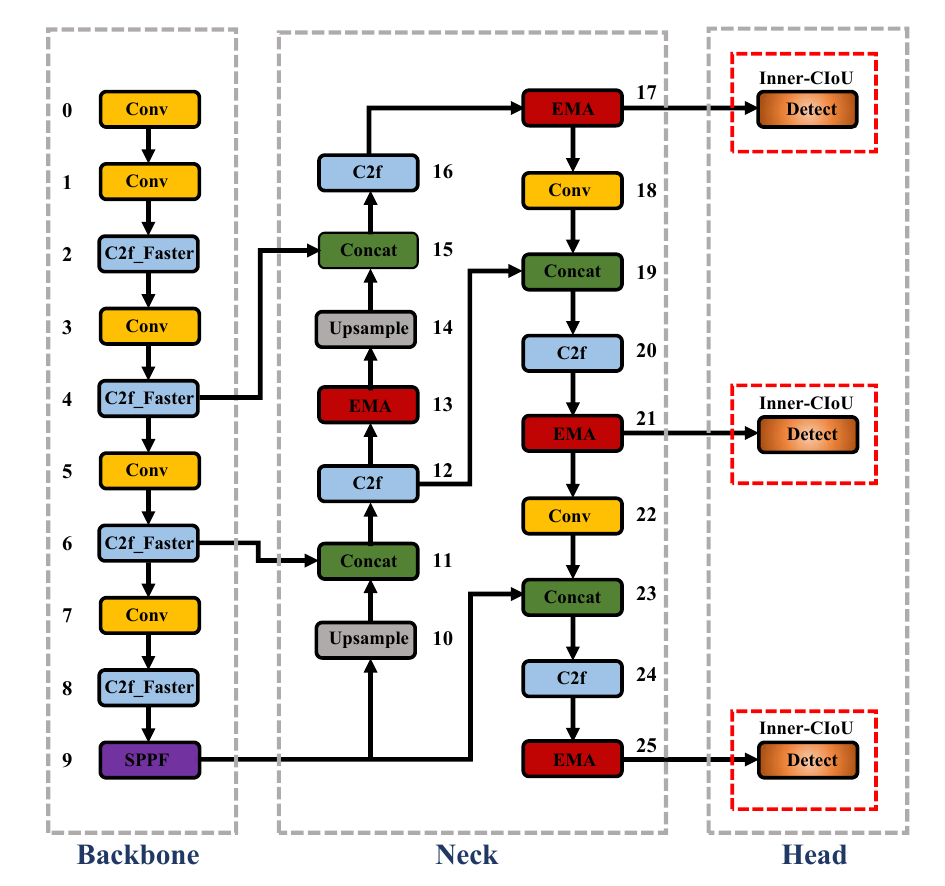}
    \caption{Architecture of the proposed AIR-YOLO network. Key enhancements include a lightweight backbone incorporating  Fasternet blocks comprising C2f\_Faster, the integration of EMA attention modules into the neck, and the adoption of the Inner-IoU loss function in the head for more accurate bounding box regression.}
    \label{fig:network}
\end{figure*}

\subsection{Object Detection}
Object detection (OD) methods are generally categorized into two types: two-stage and one-stage methods. Two-stage methods, including R-CNN~\cite{girshick2014rich}, Fast R-CNN~\cite{girshick2015fast}, and Faster R-CNN~\cite{ren2015faster}, first generate regions of interest (RoIs) and then classify and localize objects within these regions. This division allows for more precise detection but can be computationally expensive. 
In contrast, one-stage methods, such as the YOLO series~\cite{redmon2016you,bochkovskiy2020yolov4,li2022yolov6,wang2023yolov7,chen2019mmdetection, cheng2024yolo, Jocher_Ultralytics_YOLO_2023,ding2023cf,qin2022id,ning2024yolov7} and SSD~\cite{liu2016ssd}, streamline the process by integrating detection and classification into a single, unified framework, enabling real-time performance with a trade-off in accuracy.

Object detection (OD) has seen significant advancements with the introduction of transformer-based models like DETR~\cite{carion2020end}, which reconceptualizes OD as a query prediction task, offering a new perspective on object detection methodologies. 
Building on this progress, the emergence of large language models has inspired researchers to explore their application in OD tasks, as demonstrated in recent works~\cite{zang2024contextual, chen2024taskclip}. 
These developments have expanded the capabilities of OD systems, enabling them to handle more complex and diverse detection scenarios. 
However, significant challenges remain, particularly in achieving robust and consistent detection accuracy under varying environmental conditions, such as lighting changes and occlusions.

In the field of airplane surface damage detection,  detecting damage, such as cracks, dents, or scratches, requires high precision to ensure safety and reliability, while also demanding real-time processing capabilities to facilitate timely inspections. 
Traditional methods often struggle to balance these competing demands. 
To address this, we propose an improved YOLO-based detection network, specifically optimized for the unique requirements of airplane surface damage detection.

\subsection{Damage Detection}
Damage detection plays a crucial role in a wide range of industrial applications. 
Cha et al.~\cite{cha2017deep} propose a vision-based method utilizing convolutional neural networks (CNNs) for detecting concrete cracks, achieving an accuracy of 98\% without relying on traditional image processing techniques. This method demonstrates superior performance over conventional approaches, showing robustness and adaptability across varying real-world environments.
Similarly, Lin et al.~\cite{lin2017structural} develop a deep learning-based damage detection approach that automatically extracts relevant features from low-level sensor data. 
Their model demonstrates high localization accuracy, even in the presence of noisy data. 
The learned features offer valuable insights into the network’s capacity to handle structural data effectively.
Nex et al.~\cite{nex2019structural} introduce an advanced CNN for detecting visible structural damage from disaster scenarios. 
They highlight the network’s effectiveness under real-world conditions, particularly in terms of its geographical transferability across different platforms and resolutions. 
{Cui et al.~\cite{cui2023skip} present an automated inspection method for noise barriers using UAV images. 
They propose a defect detection technique based on a fully convolutional network (FCN) named skip connection YOLO Detection Network (SCYNet). 
It incorporates a skip-connected feature structure, Simi-BiFPN, which effectively fuses features from various scale layers. 
Liu et al.~\cite{liu2022uav} explore UAV-based systems for automated aircraft inspection, integrating visual–inertial odometry (VIO) with ArUco markers to enhance localization accuracy through adaptive constraints and joint optimization.
Transfer learning is employed to address limited training data for defect detection. 
These approaches enable precise fuselage defect localization by correlating UAV pose estimates from optimized VIO with high-resolution imagery, balancing marker-based constraints and detection robustness.}

Existing methods for damage detection have demonstrated considerable success, challenges persist in specific domains, particularly in detecting damage on airplane surfaces, primarily due to the limited availability of image datasets. 
To address this critical gap, we have developed the Airplane Surface Damage Dataset (AIRSD), which serves as a valuable resource to facilitate and advance research and development in the field of automated aircraft inspection.

\section{Methodology}
Upon each instance of an aircraft taxiing into the hangar, our system performs a comprehensive photographing and scanning procedure, capturing both 2D imagery and 3D point cloud data, as illustrated in Fig.~\ref{fig:pipeline}. 
The first phase focuses on damage detection within the acquired 2D images using a novel network derived from YOLOv8 \cite{Jocher_Ultralytics_YOLO_2023}. 
This architecture is optimized for lightweight and accurate identification of airplane surface damage. 
Once damage is identified in the 2D images, it is subsequently mapped onto the corresponding 3D point clouds, enabling precise 3D localization on the aircraft surface. 
This approach facilitates a more accurate and efficient detection and localization of surface damage, significantly enhancing the aircraft maintenance process by providing detailed spatial information for repair and inspection workflows.

\subsection{Improved YOLOv8 Architecture}
After obtaining a set of images of the aircraft surface, our next step is to detect the damage, such as cracks, dents, and scratches.
In this paper, we propose AIR-YOLO(Fig.\ref{fig:network}), a novel detection network derived from the YOLOv8 architecture. 
The YOLOv8 framework was selected as the baseline due to its proven balance of real-time processing capabilities and high detection accuracy, critical for large-scale aircraft inspections requiring both efficiency and precision. 
The original architecture comprises three core components: a backbone for hierarchical feature extraction, a neck for multi-scale feature aggregation, and a head for bounding box regression and classification. 
Building upon this foundation, AIR-YOLO introduces three targeted enhancements to address the challenges of aircraft surface damage detection:
(i) lightweight FasterNet blocks in the backbone to enhance computational efficiency,
(ii) EMA attention modules in the neck to prioritize multi-scale defect features, and
(iii) an Inner-CIoU loss function to refine bounding box regression by focusing on critical damage regions.
{
These modifications collectively improve accuracy while maintaining computational practicality, as detailed below.
\begin{itemize}
    \item \textbf{FasterNet blocks in the backbone}:  
    To address deployment constraints in aviation scenarios, we prioritized lightweight design. FasterNet’s Partial Convolutions (PConv) reduce redundant computations while retaining feature richness. By replacing only bottleneck layers with FasterBlocks, we achieved a 30\% parameter reduction without significant accuracy loss, ensuring suitability for resource-limited environments. 
    \item \textbf{EMA attention modules in the neck}:  
    Aircraft surfaces exhibit scale-varying defects (e.g., cracks vs. dents). EMA modules enhance multi-scale feature aggregation by combining channel and spatial attention across parallel branches. This design suppresses background noise and prioritizes critical defect regions, improving mAP by 0.9\%.   
    \item \textbf{Inner-CIoU loss function}:  
    Traditional CIoU struggles with small or elongated defects common in aircraft imagery. Inner-CIoU introduces adaptive auxiliary bounding boxes to focus regression on defect centers, accelerating convergence and improving localization precision.
\end{itemize}
}

\begin{figure}[ht]
    \centering
    \includegraphics[width=1.\linewidth]{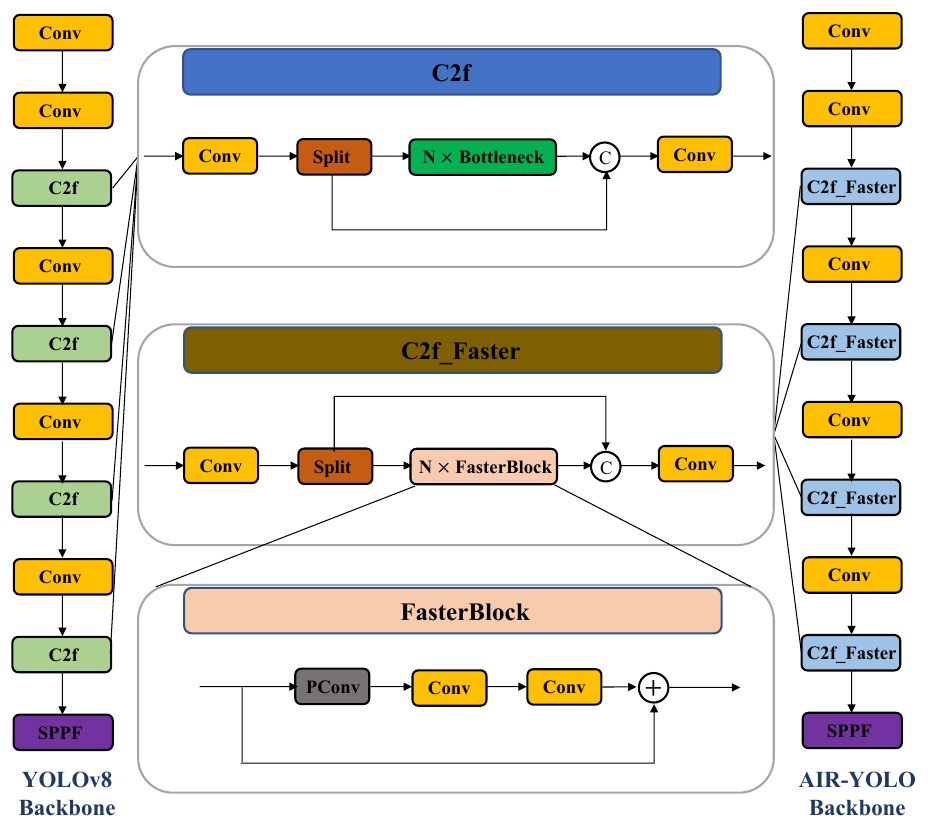}
    \caption{Backbone structure of our AIR-YOLO.}
    \label{fig:backbone}
\end{figure}

\begin{table*}[ht]
\centering
\caption{Parameter Counts of YOLOv8 and AIR-YOLO in Backbone}
\label{tab:params_comparison}
\begin{tabular}{c|cccc|cccc} \hline 
{NO.} & {Module (YOLOv8)} & {Input} & {Output} & {Params} & {Module (AIR-YOLO)} & {Input} & {Output} & {Params} \\ \hline
0     & Conv            & 3      & 16     & 464     & Conv            & 3      & 16     & 464     \\ 
1     & Conv            & 16     & 32     & 4672    & Conv            & 16     & 32     & 4672    \\ 
2     & C2f             & 32     & 32     & 7360    & C2f\_Faster     & 32     & 32     & 3920    \\ 
3     & Conv            & 32     & 64     & 18560   & Conv            & 32     & 64     & 18560   \\ 
4     & C2f             & 64     & 64     & 49664   & C2f\_Faster     & 64     & 64     &    22144 \\ 
5     & Conv            & 64     & 128    & 73984   & Conv            & 64     & 128    & 73984   \\ 
6     & C2f             & 128    & 128    & 197632  & C2f\_Faster     & 128    & 128    & 87552  \\ 
7     & Conv            & 128    & 256    & 295424  & Conv            & 128    & 256    & 295424  \\ 
8     & C2f             & 256    & 256    & 460288  & C2f\_Faster     & 256    & 256    &  240128 \\ 
9     & SPPF            & 256    & 256    & 164608  & SPPF            & 256    & 256    & 164608  \\ \hline
Sum   &                 &        &        & 1272656 &                 &        &        & 911426 \\ \hline
\end{tabular}
\end{table*}

\subsection{Backbone}

A key performance metric for detectors in airplane surface damage is lightweight design, particularly for potential on-board deployment, where optimizing both accuracy and speed with limited computational resources is essential. 
However, achieving an effective lightweight network requires a careful balance between reducing parameters and maintaining accuracy.

Fasterent~\cite{chen2023run} leverages Partial Convolution (PConv) to optimize network performance by focusing on channels that contribute most to computational efficiency, thereby increasing FLOPs in targeted areas. This approach reduces redundancy in feature maps by applying standard convolution operations selectively to a subset of input channels. 
In AIR-YOLO, the C2f\_Faster module incorporates the FasterBlock structure from FasterNet, as illustrated in Fig.\ref{fig:backbone}. 
Following the design principles in \cite{chen2023run}, 3/4 of the channels in C2f\_Faster are processed using 3$\times$3 convolutions, followed by two pointwise convolutions (1$\times$1) applied after PConv to maintain feature depth and spatial integrity.
However, FasterNet’s results indicate that replacing standard convolutions entirely with PConv results in a marked accuracy drop.
To mitigate this in AIR-YOLO, we replace only the bottleneck layers within the C2f module with FasterBlock, ensuring that crucial feature information traverses all channels with minimal compromise in accuracy. 
As shown in Table~\ref{tab:params_comparison}, this configuration reduces AIR-YOLO’s backbone parameters by around 30\% compared to YOLOv8 while maintaining high performance.
\begin{figure}[ht]
    \centering
    \includegraphics[width=1.\linewidth]{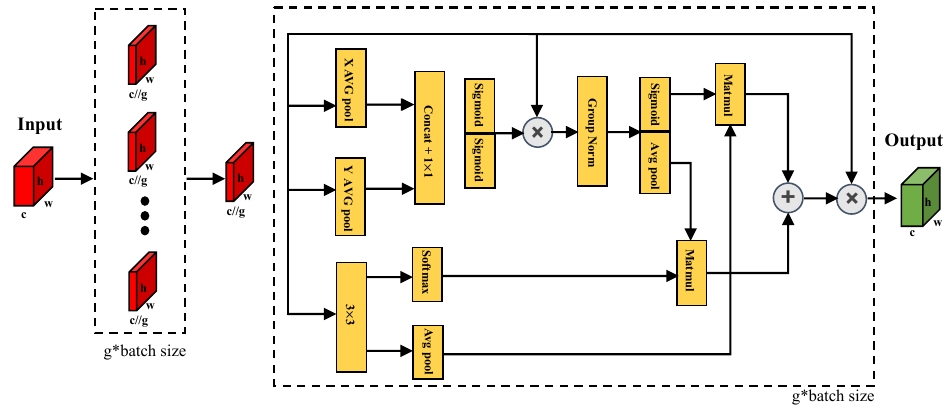}
    \caption{Structure of EMA attention module. }
    \label{fig:EMA}
\end{figure}

\subsection{Feature Aggregation}
Given that the background information in airplane surface images is inherently less complex than that of natural images, we introduce Efficient Multi-scale Attention (EMA) modules into the neck layer of our network. These modules are strategically positioned downstream of the C2f module to optimize the aggregation of multi-scale features. 
By leveraging the EMA modules, we enhance the network’s ability to capture and integrate discriminative features, thereby improving the overall representational capacity of the model. 
It is designed to address the unique characteristics of airplane surface imagery, ensuring robust feature extraction and aggregation.

EMA is an efficient attention module designed to capture multi-scale features based on cross-spatial learning~\cite{ouyang2023efficient}. As depicted in Fig.\ref{fig:EMA}, it consists of two parallel branches: a 1x1 convolutional branch and a 3x3 convolutional branch. 
The 1x1 branch performs channel attention encoding via one-dimensional global average pooling, while the 3x3 branch captures fine-grained multi-scale spatial features. 
These branches then undergo cross-space interaction learning, producing a final attention map that is multiplied by the original input feature map.
The resulting attention map not only captures the relative importance of regions within the same scale but also highlights features across different scales. 
This multi-scale attention mechanism effectively suppresses background features and enhances the network’s ability to detect multi-scale airplane surface damage. 
Consequently, the EMA module significantly improves pixel-level attention, ensuring the network remains focused on the essential target areas without reducing dimensionality, leading to more accurate damage detection.

\subsection{Loss Function}
The bounding box regression loss function plays a crucial role in object detection by quantifying the discrepancy between predicted bounding boxes and ground truth (GT) data. 
In YOLOv8, the loss function comprises three components, which are organized into two branches: classification and regression. For classification, the binary cross-entropy (BCE) loss is used, while regression tasks rely on the complete intersection over union (CIoU) loss and distribution focal loss (DFL).

However,  airplane surface images often present unique characteristics that differ from natural images. 
These differences can bring challenges to the effectiveness of standard CIoU loss, which may not fully capture the specific patterns inherent in airplane damage imagery. 
To address this issue, we employ an enhanced loss function, {Inner-CIoU}, an extension of CIoU designed to improve accuracy and accelerate convergence~\cite{zhang2023inner}. 
It leverages auxiliary bounding boxes, which are scaled versions of both the ground truth and predicted bounding boxes, offering more accurate control over the bounding box regression process.

In Inner-CIoU, a scaling factor is employed to adjust the size of these auxiliary bounding boxes.  The value is typically between the range [0.5, 1.5]. This flexibility allows for a more refined approach to bounding box regression, making it effective for airplane damage detection. 
The ground truth (GT) box and anchor are denoted as \( B_{gt} \) and \( B \), respectively, while the center points of the GT box and anchor are represented by \( (x_{c}^{gt}, y_{c}^{gt}) \) and \( (x_c, y_c) \). The width and height of the GT box are denoted as \( w^{gt} \) and \( h^{gt} \), and those of the anchor by \( w \) and \( h \).
\begin{equation}
b_{l}^{gt} = x_{c}^{gt} - \frac{w^{gt} \cdot Ratio}{2}, \quad b_{r}^{gt} = x_{c}^{gt} + \frac{w^{gt} \cdot Ratio}{2}
\end{equation}

\begin{equation}
b_{t}^{gt} = y_{c}^{gt} - \frac{h^{gt} \cdot Ratio}{2}, \quad b_{b}^{gt} = y_{c}^{gt} + \frac{h^{gt} \cdot Ratio}{2}
\end{equation}

\begin{equation}
b_l = x_c - \frac{w \cdot Ratio}{2}, \quad b_r = x_c + \frac{w \cdot Ratio}{2}
\end{equation}

\begin{equation}
b_t = y_c - \frac{h \cdot Ratio}{2}, \quad b_b = y_c + \frac{h \cdot Ratio}{2}
\end{equation}

\begin{equation}
\begin{aligned}
inter &= ( \min(b_r^{gt}, b_r) - \max(b_l^{gt}, b_l)) \cdot \\
&\quad ( \min(b_b^{gt}, b_b) - \max(b_t^{gt}, b_t) )
\end{aligned}
\end{equation}

\begin{equation}
union = (w^{gt} \cdot h^{gt}) \cdot (Ratio)^2 + (w \cdot h) \cdot (Ratio)^2 - inter
\end{equation}

\begin{equation}
IoU_{inner} = \frac{inter}{union}
\end{equation}

Finally, the Inner-CIoU loss is computed as:

\begin{equation}
L_{{Inner-CIoU}} = L_{{CIoU}} + IoU - IoU_{{inner}},
\end{equation}

The use of auxiliary bounding boxes allows the loss function to accelerate convergence for high IoU samples through smaller Ratios, while larger Ratios improve the regression for low IoU samples. This adaptive approach enhances both detection accuracy and robustness, particularly in the field of damage detection in airplane imagery.

\subsection{Damage Localization}
Upon detecting the surface damage in 2D images, the next step involves accurately localizing the corresponding  anomaly regions within 3D point clouds. 
Using the intrinsic and extrinsic parameters of the camera, we compute the 3D positions of the identified damage. 
This process enables the projection of image pixels onto the 3D point clouds, thereby achieving an effective mapping between the 2D image space and the 3D model, as outlined in Algorithm~\ref{alg:localization}.
As a result, the defect locations are determined, providing essential guidance for human operators during the airplane inspection process.

\begin{algorithm}
\caption{  Damage Localization}\label{alg:localization}
\begin{algorithmic}

\State \textbf{Input:}
\State $B = \{(x_{\text{min}}, y_{\text{min}}), (x_{\text{max}}, y_{\text{max}})\}$: Bounding box of damage in the 2D images
\State $K$: Camera intrinsic parameters matrix 
\State $R$: Rotation matrix from camera's extrinsic parameters 
\State $T$: Translation vector from camera's extrinsic parameters
\State $P_{3D}$: 3D point clouds from the laser scanner

\State \textbf{Output:}
\State $P_{3D\_damaged}$:  Damage points in the point cloud 

\Statex
\Procedure {Localize Damage In 3D Point Clouds}{}
    \State \textbf{Compute projection matrix:}
    \State $P \gets K \cdot [R \ | \ T]$ \Comment{Compute projection matrix}

    \State \textbf{Project 3D points onto the 2D image plane:}
    \State $P_{2D\_from\_3D} \gets \{\}$ \Comment{Initialization}
    \For{each $(X_i, Y_i, Z_i) \in P_{3D}$}
        \State $\begin{pmatrix} u_i \\ v_i \\ 1 \end{pmatrix} \gets P \cdot \begin{pmatrix} X_i \\ Y_i \\ Z_i \\ 1 \end{pmatrix}$
        \State Add $(u_i, v_i, Z_i)$ to $P_{2D\_from\_3D}$
    \EndFor

    \State \textbf{Determine the range of 2D points in the bounding box:}
    \State $P_{2D} \gets \{(x, y, Z) \ | \ x_{\text{min}} \leq x \leq x_{\text{max}}, y_{\text{min}} \leq y \leq y_{\text{max}}, (x, y, Z) \in P_{2D\_from\_3D}\}$

    \State \textbf{Back-project 2D points to 3D space using depth:}
    \State $P_{3D\_damaged} \gets \{\}$ \Comment{Initialiation of localized points}
    \For{each $(x, y, Z) \in P_{2D}$}
        \State $\begin{pmatrix} X \\ Y \\ Z \\ 1 \end{pmatrix} \gets P^{-1} \cdot \begin{pmatrix} x \cdot Z \\ y \cdot Z \\ Z \\ 1 \end{pmatrix}$
        \State Add $(X, Y, Z)$ to $P_{3D\_damaged}$
    \EndFor
    
    \State \textbf{Output} $P_{3D\_damaged}$
\EndProcedure
\end{algorithmic}
\end{algorithm}

\section{Results and discussion}
\label{experiment}
In this section, we verify the performance of the proposed framework. 
We first provide a detailed description of the constructed dataset, along with the training configurations. 
It is followed by a discussion of the evaluation metrics selected to assess the model’s performance. 
To further analyze the contribution of individual components within the framework, we conduct ablation studies, examining the impact of different modules. Subsequently, we present a comparative analysis of our proposed network against several state-of-the-art object detection methods, highlighting its effectiveness. 
Finally, we report the results of both 2D damage detection and 3D localization to demonstrate the practical capabilities of our framework.

% \begin{figure}[htbp]
%     \centering
%     \includegraphics[width=0.95\linewidth]{images/dataset.png}
%     \caption{The label distribution in our AIRSD dataset. }
%     \label{fig:dataset}
% \end{figure}
\subsection{Dataset and Training} 
We introduce the airplane image repository for surface damage (AIRSD), a novel and comprehensive dataset specifically designed for the detection and analysis of airplane damage. 
The dataset comprises 8,091 images, encompassing 11 distinct types of damage: crack, dent, rust, paint peeling, scratch, rivet damage, lightning strike, bird strike, hail damage, wrinkle, and missing fastener. 
To ensure robust and diverse coverage of damage scenarios, AIRSD integrates both synthetic and real-world data. The synthetic data is generated through a systematic process of overlaying damage masks onto clean airplane surface backgrounds, simulating a wide range of damage conditions under controlled parameters. Notably, the test set is exclusively composed of real-world images, ensuring the evaluation of models under realistic and practical conditions.

All damage regions within the dataset have been meticulously annotated and labeled using X-AnyLabeling~\cite{wang2023advanced}, a state-of-the-art annotation tool, to guarantee precision and reliability. 
To facilitate compatibility with a broad spectrum of deep learning frameworks, annotations are provided in both COCO and YOLO formats. 
The dataset reveals a non-uniform distribution of damage types, with certain categories, such as paint peeling, occurring at significantly higher frequencies compared to rarer instances like bird strikes. 
It serves as a foundational resource for advancing research in airplane damage detection, offering a balanced combination of synthetic and real-world data to address the challenges of variability and generalization in this domain.

Our AIR-YOLO is implemented in PyTorch, with an NVIDIA RTX3090 GPU and CUDA11.8 environment for both training and inference tests.
The Adam optimizer is used for network parameter updates and optimization, and the initial learning rate for weight is set to 0.001.
The models have a batch size of 16 and are trained for 400 epochs.
\begin{table*}[ht]
\centering
\caption{Results of Ablation experiments on the dataset}
\label{tab:ablation}
\begin{tabular}{cccc|cccc}
\hline
\multicolumn{4}{c|}{METHOD}                                                                                   & \multirow{2}{*}{P (\%)} & \multirow{2}{*}{R (\%)} & \multirow{2}{*}{F1 (\%)} & \multirow{2}{*}{mAP (\%)} \\ \cline{1-4}
YOLOv8                    & Lightweight backbone                   & EMA                   & Inner-CIoU                    &                  &                   &                     &                     \\ \hline
\checkmark  &                           &                           &                           &         58.8          &         \textbf{67.6}          &    62.8           &     62.4                \\
 \checkmark & \checkmark &                           &                           &        60.6            &              61.5       &          61.0          &           61.8          \\
 \checkmark &  &  \checkmark &  &         68.7          &         62.1           &      65.2         &         63.3           \\
  \checkmark &  &   & \checkmark  &          70.4         &      62.2             &      66.9           &     62.7                \\
    \checkmark & \checkmark &  \checkmark & \checkmark  &         \textbf{71.3}            &     64.4             &     \textbf{67.7}       &        \textbf{65.2}               \\ \hline
\end{tabular}
\end{table*}
\subsection{Evaluation Metric}
We use the Precision (\textit{P}), Recall (\textit{R}), F1, and mAP scores to evaluate the performance of different methods, computed as,
\begin{equation}
    P = \frac{\text{TP}}{\text{TP} + \text{FP}} 
\end{equation}

\begin{equation}
    R = \frac{\text{TP}}{\text{TP} + \text{FN}}
\end{equation}

\begin{equation}
    F_1= 2     \times       \frac{P\times R}{P+R}
\end{equation}

\begin{equation}
    AP= \int_0^1 p(r) dr
\end{equation}

\begin{equation}
    \text{mAP} = \frac{1}{N} \sum_i AP_i
\end{equation}

{TP represents the number of correctly identified airplane defects, FP denotes the number of falsely identified defects, and FN refers to the number of missed defects. 
Precision (P) and Recall (R) are critical for safety-critical applications like aircraft inspection. 
High precision minimizes false positives (avoiding unnecessary repairs), while high recall ensures minimal missed defects (preventing safety risks).
F1 Score balances precision and recall, addressing class imbalance in the AIRSD dataset (e.g., rare damage types like bird strikes).
mAP@0.5 (mean Average Precision at IoU=0.5) is standard in object detection benchmarks. A lower IoU threshold is pragmatic for initial damage detection, as small defects (e.g., cracks) may occupy limited pixel areas.
FPS (Frames Per Second) underscores our model’s real-time applicability, crucial for rapid inspections.
These metrics collectively ensure robustness, comparability with state-of-the-art methods, and alignment with aviation safety requirements.}

\subsection{Ablation Study}
We conduct an ablation study on the proposed AIR-YOLO model, evaluating the impact of key improvements, including an improved backbone, the integration of EMA modules, and modifications to the loss function. 
These experiments assess the contributions of each component to the detection accuracy. 
Quantitative results from these experiments are presented in Table~\ref{tab:ablation}, with YOLOv8~\cite{Jocher_Ultralytics_YOLO_2023} serving as the baseline model for comparison.

\textbf{Lightweight Backbone.}  
The lightweight backbone architecture, which incorporates Fasternet blocks, significantly reduces the number of model parameters compared to the baseline YOLOv8. While this reduction in complexity results in a marginal decrease of 0.6\% in mean Average Precision (mAP), it substantially improves computational efficiency. 
This trade-off demonstrates that the lightweight backbone achieves a balance between model simplicity and detection accuracy, making it suitable for resource-constrained applications.

\textbf{EMA Module.}  
The inclusion of EMA attention modules enhances both precision and recall, as evidenced by the results in Table~\ref{tab:ablation}. 
Specifically, the EMA module contributes to a 0.9\% increase in mAP, underscoring its effectiveness in directing the model’s attention to relevant regions of the input. This improvement highlights the module’s ability to refine feature representations, leading to more accurate object detection and classification.

\textbf{Inner-CIoU Loss Function.}  
The Inner-CIoU loss function introduces auxiliary bounding boxes to improve detection performance.
By focusing on the central regions of bounding boxes, this approach increases Inner-IoU values, which play a more significant role in the overall loss computation. 
This mechanism accelerates the regression process of predicted bounding boxes, ensuring better alignment with ground truth annotations. 
The emphasis on central regions also enables a more precise evaluation of overlapping areas, making the Inner-CIoU loss particularly effective for detecting subtle features, such as airplane damage.

{ As demonstrated in Table~\ref{tab:iou_comparison}, the proposed Inner-CIoU loss function exhibits superior performance in damage detection accuracy compared to other IoU-based loss functions. The results indicate that Inner-CIoU achieves an mAP of 65.2\%, outperforming Giou (62.9\%), CIoU (62.4\%) and DIoU (61.7\%) by 2.3\%, 2.8\% and 3.5\%, respectively. Additionally, Inner-CIoU attains the highest precision (71.3\%) and F1 score (67.7\%), highlighting its robustness in balancing detection accuracy and localization precision.
The performance improvement validate the effectiveness of Inner-CIoU in addressing the challenges of aircraft damage detection. }
\begin{table}[h]
\caption{Performance Comparison of Different IoU Loss Functions}
\label{tab:iou_comparison}
\centering
\begin{tabular}{lrrrr}
\toprule
\textbf{Loss Function} & \textbf{P} & \textbf{R} & \textbf{F1} & \textbf{mAP} \\
\midrule
GIoU          & 60.1    & 68.3      & 63.8   & 62.9   \\
DIoU          & 61.8    & \textbf{69.5}    & 65.3   & 61.7   \\
CIoU          & 58.8    &{67.6}    & 62.8   & 62.4   \\
Inner-CIoU (Ours) & \textbf{71.3} & {64.4} & \textbf{67.7} & \textbf{65.2} \\
\bottomrule
\end{tabular}
\end{table}

\begin{figure*}[htbp]
    \centering
    \includegraphics[width=1.\linewidth]{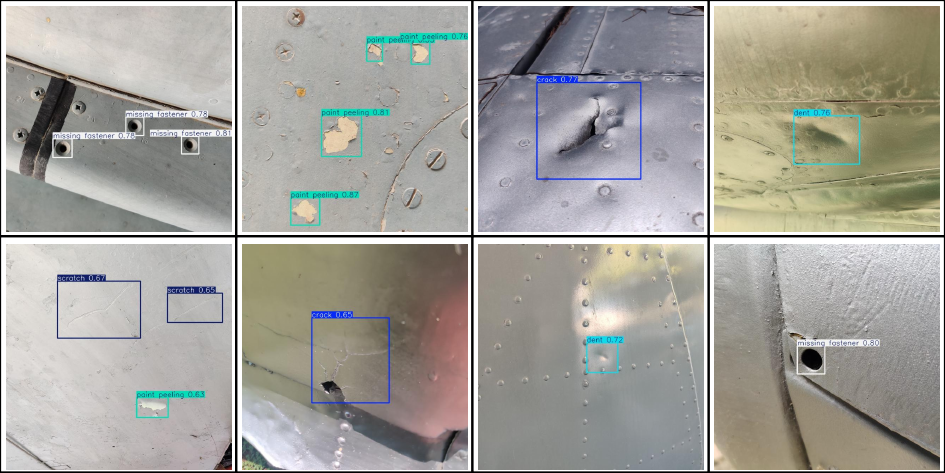}
    \caption{Damage detection results for airplane surface. }
    \label{fig:damage}
\end{figure*}
\begin{figure*}[htbp]
    \centering
    \includegraphics[width=0.9\linewidth]{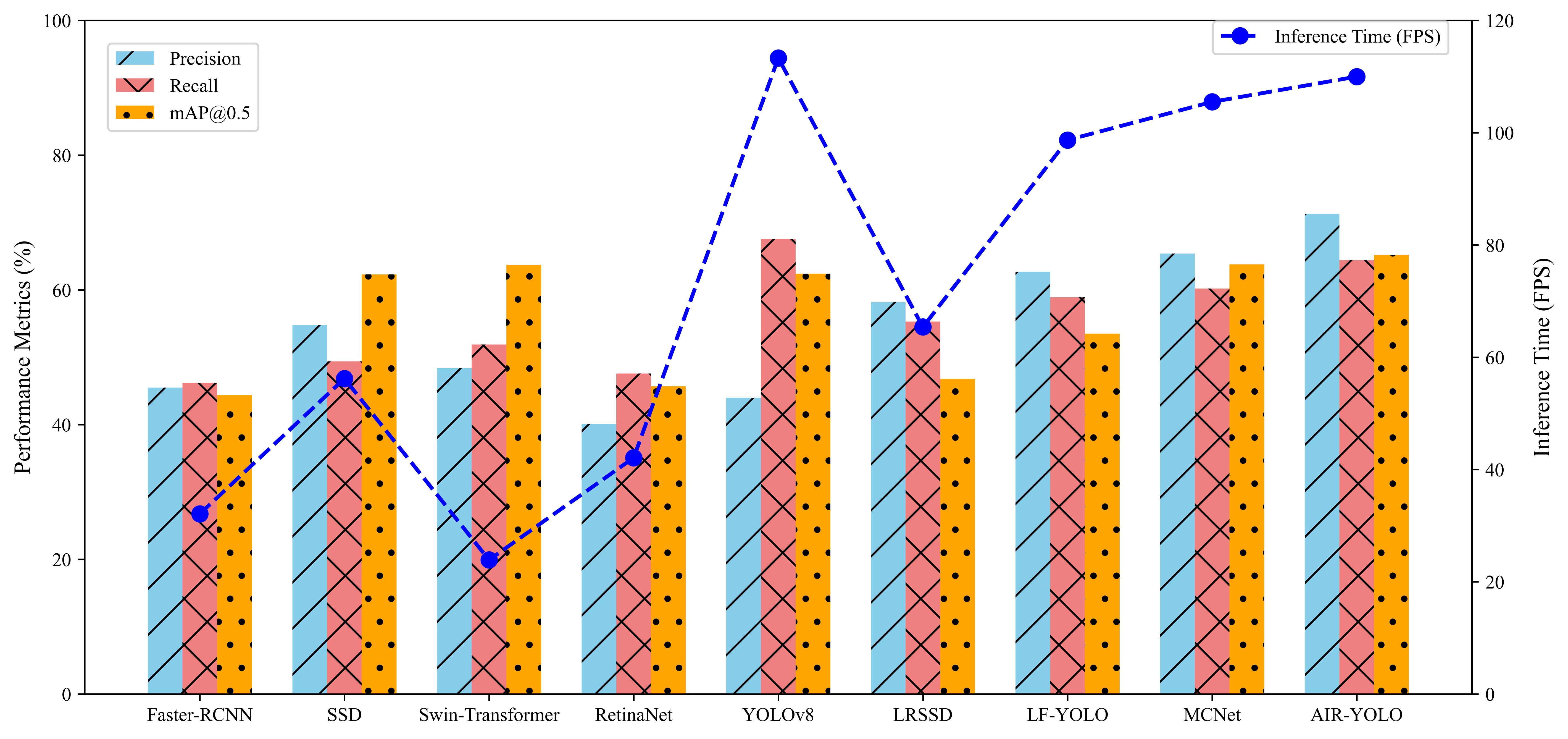}
    \caption{Performance comparison between AIR-YOLO and other SOTA models. }
    \label{fig:comparison_sota}
\end{figure*}

\begin{figure}[htbp]
    \centering
    \includegraphics[width=1.\linewidth]{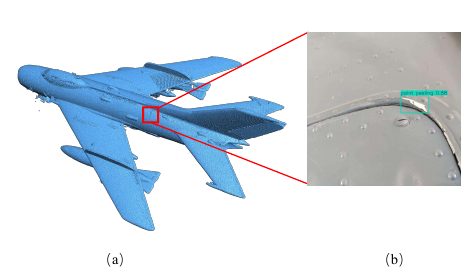}
    \caption{3D localization of damage on the airplane surface. (a) 3D point cloud of an airplane (b) one damage sample (paint peeling).}
    \label{fig:localization}
\end{figure}

\subsection{Comparison of Different Models}
To assess the reliability and effectiveness of the proposed AIR-YOLO model, we conducted a comprehensive performance comparison against several state-of-the-art object detection models using the AIRSD dataset. 
{To ensure a rigorous evaluation, we compare AIR-YOLO against both general object detectors (Faster R-CNN~\cite{ren2015faster}, SSD~\cite{liu2016ssd}, RetinaNet~\cite{lin2017focal}, Swin-Transformer~\cite{liu2021Swin}, and YOLOv8~\cite{Jocher_Ultralytics_YOLO_2023}) and state-of-the-art defect detection models (LRSSD~\cite{CANYANG2025115922}, LF-YOLO~\cite{liu2023lf} and MCNet~\cite{zhang2021_tim}).}
Each model was retrained and fine-tuned specifically for the task of airplane surface damage detection to ensure a fair comparison.
The evaluation metrics used in this study include FP, FN, P, R, mAP, and FPS, providing a balanced assessment of both accuracy and computational efficiency.

As illustrated in Fig.~\ref{fig:comparison_sota}, the AIR-YOLO model demonstrates superior performance in terms of detection accuracy, achieving an mAP of 65.2\%. 
This performance meets the stringent requirements for airplane surface damage detection, highlighting the model’s robustness in identifying subtle and complex damage patterns. 
Furthermore, AIR-YOLO exhibits a significant advantage in computational efficiency. Compared to Faster R-CNN, SSD, Swin-Transformer, and RetinaNet, AIR-YOLO achieves comparable or higher accuracy while delivering a substantially higher FPS. 
This efficiency underscores AIR-YOLO’s ability to balance high detection performance with reduced computational overhead, making it particularly suitable for real-time applications in the aviation industry.

{As illustrated in Fig.~\ref{fig:comparison_sota}, AIR-YOLO demonstrates superior detection accuracy compared to state-of-the-art defect detection methods. 
While baseline methods such as LRSSD~\cite{CANYANG2025115922} and LF-YOLO~\cite{liu2023lf}  exhibit strong performance in their respective domains (e.g., rail damage or weld inspection), their architectures lack specialized optimizations for aircraft surface defects, resulting in reduced generalization capabilities for aviation-specific scenarios. 
In contrast, AIR-YOLO’s design—featuring multi-scale EMA attention modules and the Inner-CIoU loss—explicitly addresses the unique challenges of aircraft damage detection. 
This domain-specific optimization enables a 18.4\% and 11.7\% improvement in mAP over LRSSD and LF-YOLO, respectively, while maintaining real-time efficiency (110 FPS). 
These results validate that tailored architectural innovations are critical for achieving robust performance in aviation safety applications.}

\subsection{Discussion}
\begin{table}[h]
\centering
\caption{Generalization performance of AIR-YOLO across different damage types.}
\label{tab:cross_type}
\begin{tabular}{lcccc}
\toprule
\textbf{Damage Type} & \textbf{P} & \textbf{R} & \textbf{F1} & \textbf{mAP} \\
\midrule
Crack & 51.8 & 59.5 & 55.4 & 53.6 \\
Dent & 61.3 & 63.2 & 62.2 & 57.7 \\
Rust & 63.2 & 67.2 & 65.1 & 69.1 \\
Paint peeling & 24.5 & 62.2 & 35.2 & 56.0 \\
Scratch & 18.4 & 40.0 & 25.2 & 39.5 \\
Rivet damage & 73.2 & 65.1 & 68.9 & 71.6 \\
Lightning strike & 45.5 & 100.0 & 62.5 & 99.5 \\
Bird strike & 56.2 & 59.7 & 57.9 & 61.3 \\
Hail damage & 68.1 & 59.1 & 63.3 & 68.1 \\
Wrinkle & 56.5 & 67.2 & 61.4 & 62.7 \\
Missing fastener & 62.6 & 80.6 & 70.5 & 82.1 \\
\midrule
All & 71.3 & 64.4 & 67.7 & 65.2\\
\bottomrule
\end{tabular}
\end{table}

{As shown in Table~\ref{tab:cross_type},  AIR-YOLO achieves robust performance across most damage types, with particularly high precision and recall for categories such as rust (mAP: 69.1\%), rivet damage (mAP: 71.6\%), and missing fastener (mAP: 82.1\%). However, we observe lower performance for scratches (mAP: 39.5\%) and paint peeling (mAP: 56.0\%), likely due to their subtle visual characteristics and variability in appearance. These findings highlight the model's strengths while also identifying areas for future improvement, particularly in detecting fine-grained or less conspicuous defects.  }

As illustrated in Fig.\ref{fig:damage}, the proposed method demonstrates a high level of accuracy in identifying defect regions and classifying their types. Moreover, the approach is capable of detecting multiple defect instances simultaneously. 
For example, in the bottom left subfigure, both of the paint peeling region and scratches are successfully identified, showcasing the method's capacity for handling complex, overlapping damage scenarios. 
It highlights the robustness of our detection algorithm in accurately recognizing and distinguishing between different forms of surface damage commonly encountered on aircraft. 
The overall results confirm that the method exhibits strong performance in detecting a wide range of surface defects, ensuring comprehensive and reliable damage assessment.

We also localize and highlight the 3D points corresponding to the damaged region, as shown in Fig.\ref{fig:localization}.
Our approach utilizes a non-contact methodology, enabling accurate and efficient damage detection and localization without risking any physical harm to the aircraft’s surface. 
This is achieved through the integration of multiple  high-resolution cameras and laser scanners, strategically positioned to capture detailed surface data.
Unlike existing UAV-based methods~\cite{liu2022uav}, which carry the risk of accidental collisions or falls that could potentially damage the aircraft, our system operates independently of such mechanisms, ensuring both the safety of the inspection process and the integrity of the aircraft. The high-precision data generated by our system allows inspectors to promptly identify and address damaged areas, significantly improving the efficiency and effectiveness of the inspection and maintenance process.

\subsection{Limitations}
Our system is based on visual defect detection, and the major limitation is that some small defects are too slight to be recognized from the images (even for human eyes), due to the limited resolution.
A possible solution is to use a better camera to collect higher-resolution images.

{While AIRSD integrates synthetic data to expand damage scenario coverage, synthetic images may not fully replicate real-world variations in lighting, surface textures, or environmental conditions. This gap could lead to reduced model generalization on purely real-world test data. To mitigate this, future work will explore domain adaptation techniques (e.g., adversarial training) and refine synthetic data generation pipelines to better mimic real-world conditions.  
The dataset exhibits a non-uniform distribution of damage types, with rare classes (e.g., bird strikes) underrepresented compared to frequent ones (e.g., paint peeling). This imbalance may bias the model toward the majority classes.  
Weighted loss functions~\cite{fernando2021dynamically} might prioritize underrepresented classes during training.}

{
The system’s performance may be affected by environmental factors such as highly reflective aircraft surfaces (e.g., polished fuselage) and dynamic lighting conditions (e.g., shadows, glare). 
These factors can introduce noise in both 2D images and 3D point clouds, potentially impacting detection accuracy. To mitigate this, future iterations could integrate polarization filters to reduce reflections, adaptive exposure control for cameras, and post-processing algorithms to normalize lighting variations in captured data.
}
\begin{figure*}[ht]
    \centering
    \includegraphics[width=1.0\linewidth]{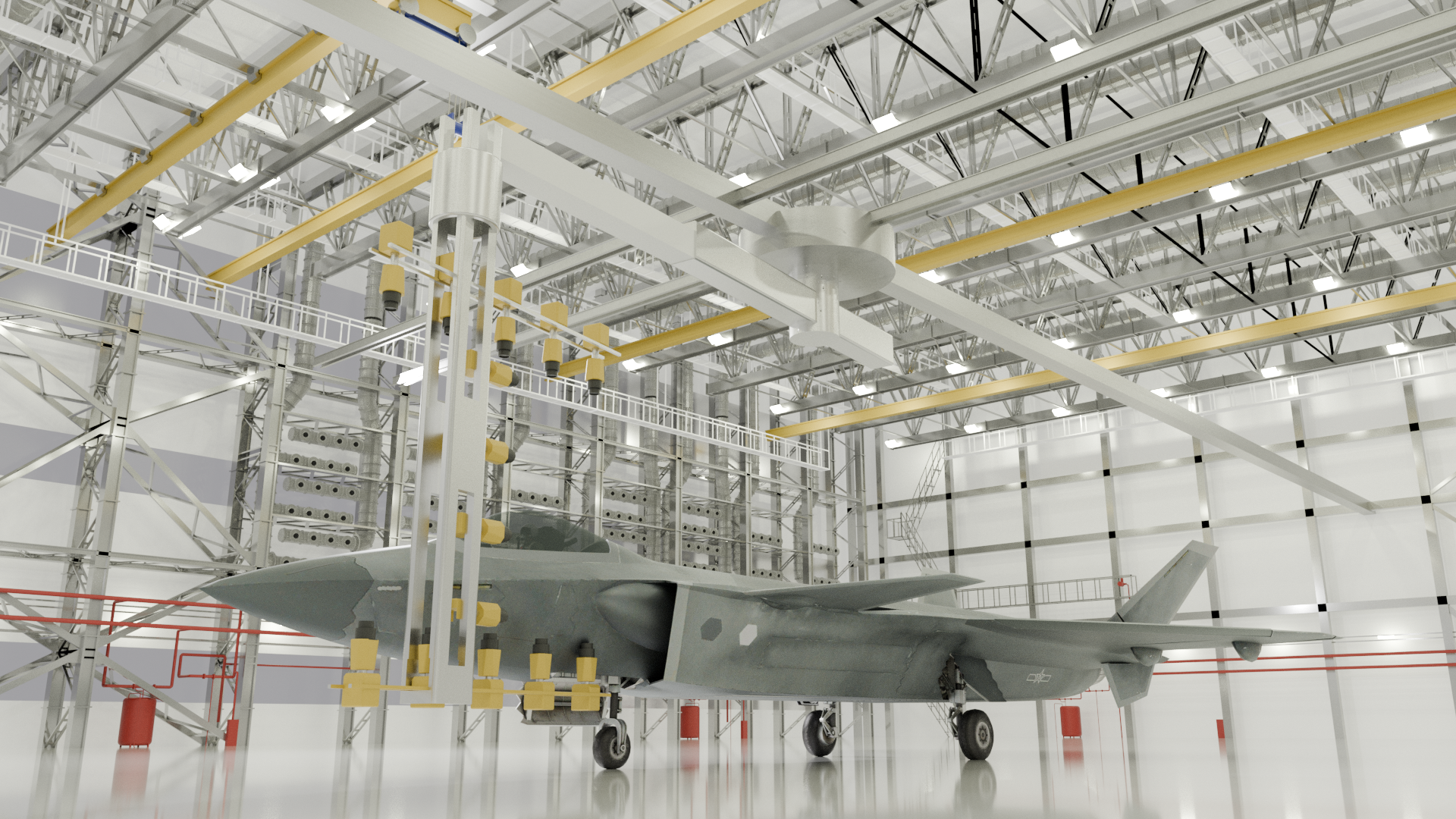}
    \caption{Our hardware platform. Cameras and scanners are mounted on the mechanical supports, enabling the acquisition of both 2D images and 3D point clouds of the airplane. }
    \label{fig:platform}
\end{figure*}

\section{Application}
Our proposed framework is implemented on an advanced inspection device, which integrates both pairs of laser scanners and high-resolution cameras, as shown in Fig.\ref{fig:platform}. 
This dual-sensor system enables the capture of both 2D and 3D data, providing high-fidelity images and corresponding point clouds that represent the airplane surface. 
By leveraging these data, the framework facilitates more comprehensive analysis, enabling accurate damage detection and 3D localization for aircraft inspection.

\subsection{Hardware platform}
The hardware platform of this device is designed for optimized mobility, particularly suited for applications involving the inspection of large, complex structures like aircraft. It operates on a rail system that is specifically engineered to align with the contours of the target aircraft, allowing for precise navigation around the fuselage and wings. This adaptive movement capability ensures complete coverage of the aircraft's surface, facilitating the acquisition of comprehensive data from all angles.
The device is equipped with a series of cameras and laser scanners strategically placed at different heights and orientations. These sensors capture both 2D high-resolution images and 3D point clouds, offering a detailed visual and spatial representation of the aircraft, including difficult-to-reach areas such as the upper fuselage and underside. 

Once the imaging and scanning process is completed, the data consisting of 2D imagery and 3D point clouds serves as the foundation for further analysis, including automated damage detection. 
By leveraging this integrated platform, the inspection process not only becomes more thorough but also significantly more efficient, reducing the likelihood of undetected damage and improving overall safety outcomes in aviation maintenance.
\begin{table}[!h]
\centering
\begin{tabular}{c c c} 
 \hline
 Method & Number of detected defects & Time \\ [0.5ex] 
 \hline
 Manual inspection & 31  &  3.1 hours\\ 
 Our method & 87  &  28 minutes\\
 \hline
\end{tabular}
\caption{Comparison of our method with manual inspection method.}
\label{tab:statistics}
\end{table}
\subsection{Comparison with manual inspection method}
To assess the performance of the proposed framework, a comparison is made with traditional human inspection methods. 
An experienced aircraft inspector was tasked with conducting a detailed manual inspection. 
As demonstrated in Table~\ref{tab:statistics}, our method substantially outperforms manual inspection in terms of speed. 
It is important to note that the total time for our approach includes the data collection, the damage detection and localization stages. 
In addition to significantly enhancing the efficiency of the inspection process, our framework also improves accuracy, offering a more reliable and scalable solution for aircraft maintenance.

\section{Conclusion}
\label{conclusion}
We present a novel framework for airplane surface damage detection and 3D localization, leveraging both image data and point clouds captured by our proposed system. 
It incorporates an enhanced deep learning network for detecting surface damage from a series of 2D images. 
This network features a lightweight backbone architecture, optimized by integrating partial convolution layers to efficiently handle data, and enhanced by the addition of EMA  attention modules in the neck of the model, which improve feature aggregation across different scales. 
Furthermore, we introduce a novel loss function, Inner-CIoU, specifically designed to enhance detection accuracy. 
Following the 2D damage detection, we extend the system’s capabilities to 3D by localizing the identified damage regions within the point clouds representing the entire airplane. 
{Experimental results demonstrate that our framework achieves a mAP of {65.2\%} at {110 FPS}, significantly improving both inspection precision and speed compared to existing manual methods.}


\bibliographystyle{IEEEtran}
\bibliography{sources/references.bib}

\begin{IEEEbiography}[{\includegraphics[width=1in,height=1.25in,clip,keepaspectratio]{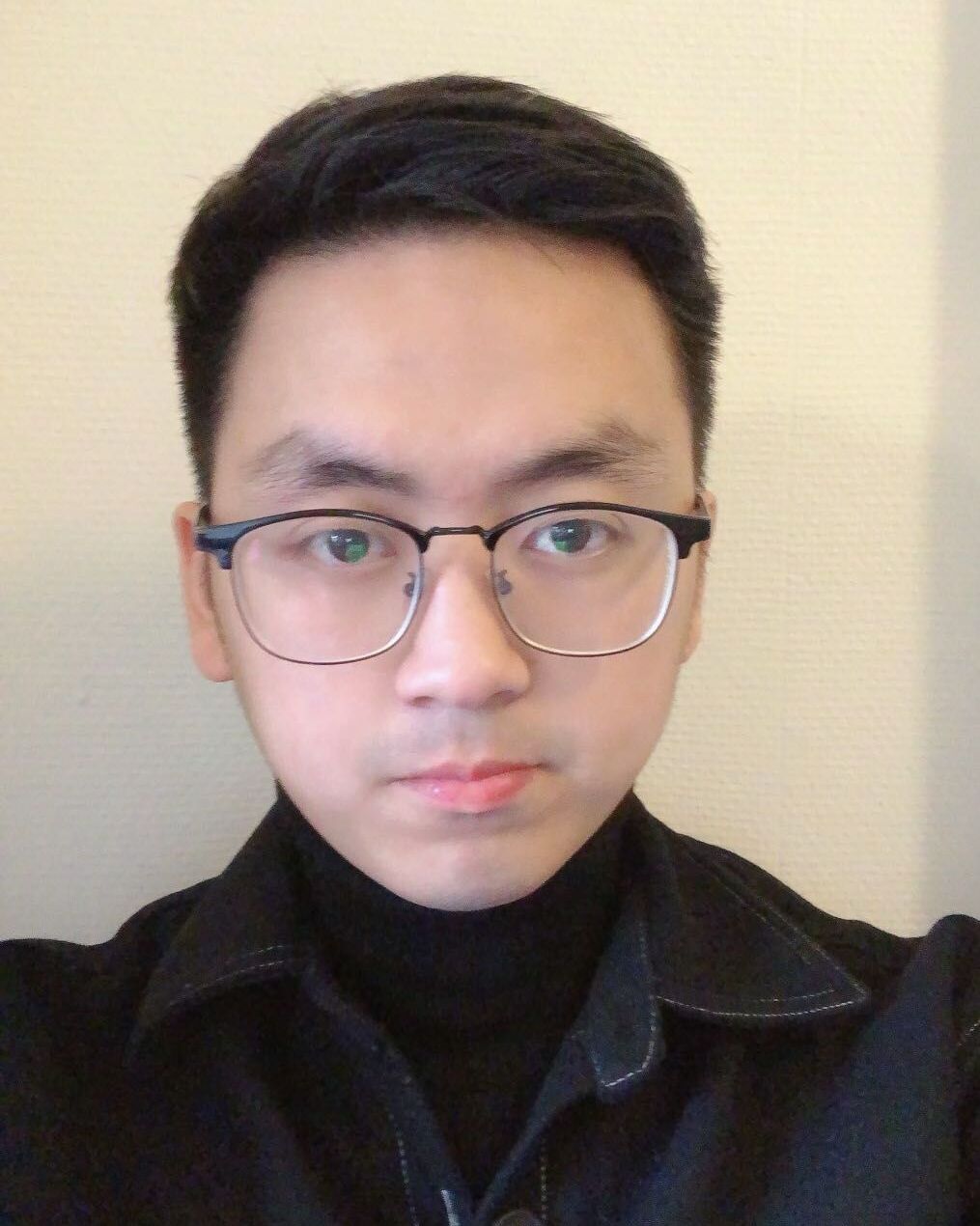}}]{Jin Huang} obtained  Ph.D. degree from Delft University of Technology (TU Delft), the Netherlands.
He received the B.S. and MSc degrees from Nanjing University of Aeronautics and Astronautics (NUAA), Nanjing, China, in 2016 and 2019, respectively. 
His research focuses on computer graphics and 3D geoinformation.
\end{IEEEbiography}

\begin{IEEEbiography}[{\includegraphics[width=1in,height=1.25in,clip,keepaspectratio]{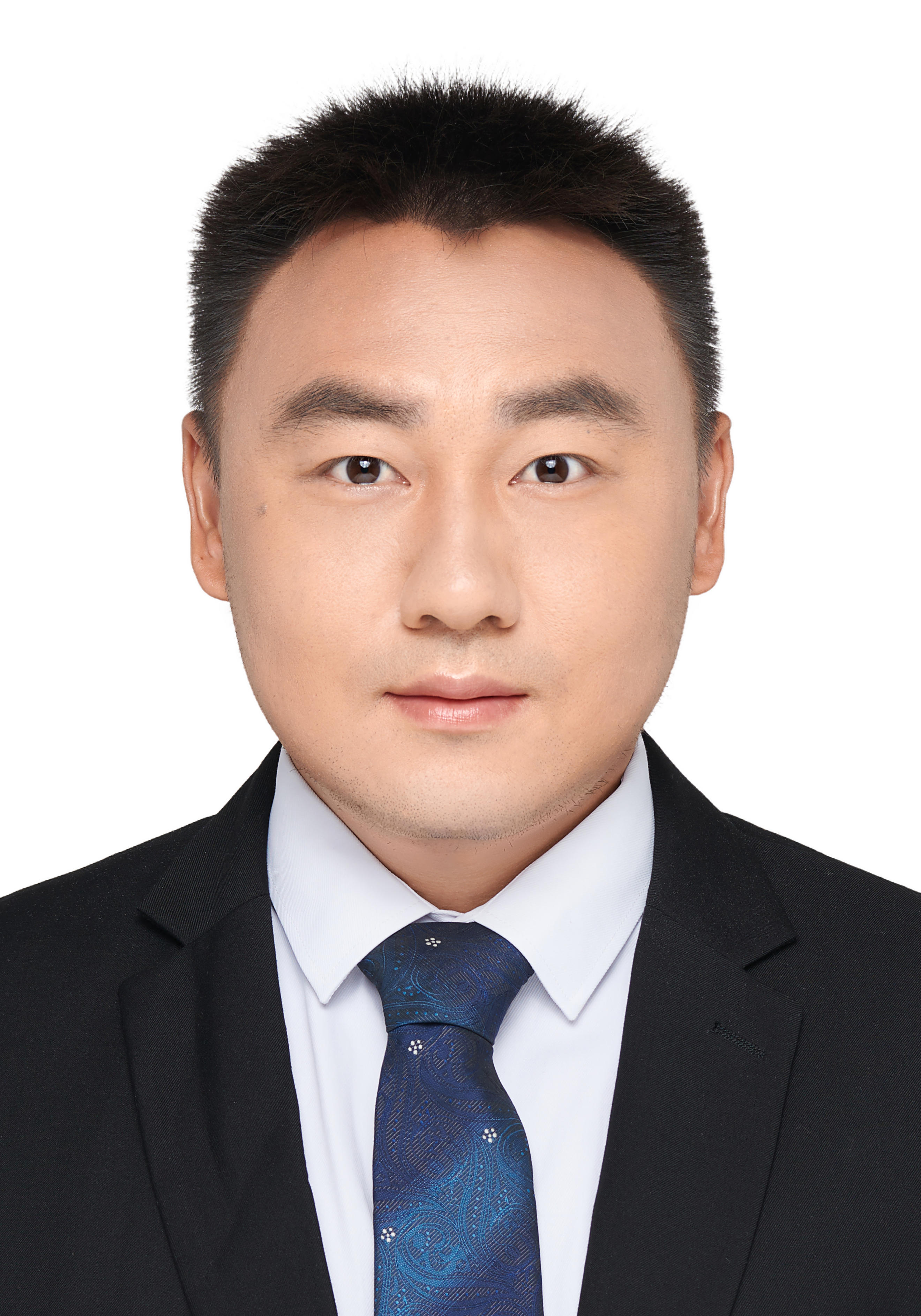}}]{Mingqiang Wei} (Senior Member, IEEE) received Ph.D. degree in computer science and engineering from The Chinese University of Hong Kong (CUHK) in 2014. He is currently a Full Professor with the School of Computer Science and Technology, Nanjing University of Aeronautics and Astronautics (NUAA). He was a recipient of the CUHK Young Scholar Thesis Awards in 2014. He is an Associate Editor of IEEE Transactions on Image Processing, and ACM TOMM; and was a leading Guest Editor of IEEE Transactions on Multimedia. He has published 150 research publications, including IEEE TRANSACTIONS ON PATTERN ANALYSIS AND MACHINE INTELLIGENCE (IEEE TPAMI), SIGGRAPH, TVCG, CVPR, and ICCV. His research interests focus on 3D vision, computer graphics, and deep learning.
\end{IEEEbiography}

\begin{IEEEbiography}[{\includegraphics[width=1in,height=1.25in,clip,keepaspectratio]{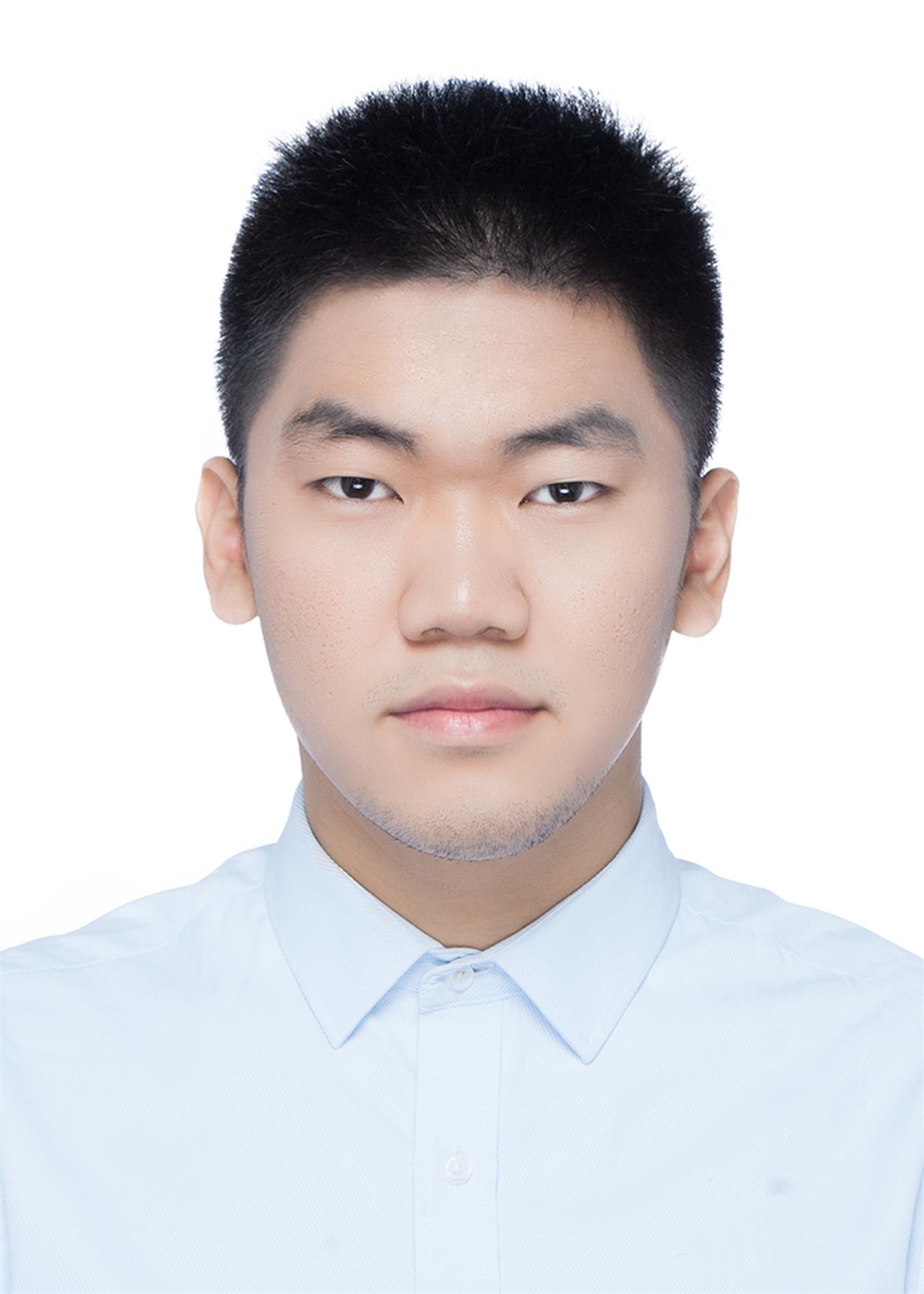}}] {Zikuan Li} received his bachelor's degree in surveying and mapping engineering and the master's degree in photogrammetry and remote sensing from Hohai University, Nanjing, China, in 2018 and 2021, respectively. He is currently pursuing the Ph.D. degree with the Nanjing University of Aeronautics and Astronautics (NUAA), Nanjing. His research interests include point cloud processing, computer vision, and deep learning.

\end{IEEEbiography}

\begin{IEEEbiography}[{\includegraphics[width=1in,height=1.25in,clip,keepaspectratio]{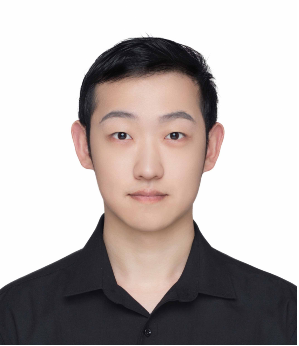}}]{Hangyu Qu} received the B.S.  degree from Shandong University of Art \& Design , Jinan, China 2023. He is currently working toward his MSc degree at Nanjing Insititute of Technology (NJIT), Nanjing, China. His research focuses on Digital Modeling and Intelligent Equipment Design.

\end{IEEEbiography}

\begin{IEEEbiography}[{\includegraphics[width=1in,height=1.25in,clip,keepaspectratio]{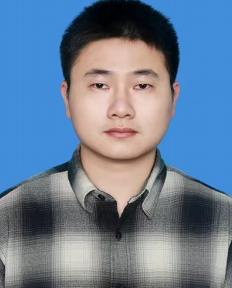}}]{Wei Zhao} received his Ph.D degree (2024) in Biological Science \& Medical Engineering from
the Southeast University, Nanjing, China. 
He is a postdoctoral fellow at the School of Computer
Science and Technology, Nanjing University of
Aeronautics and Astronautics. His research interests focus on machine learning, genetic algorithms and 3D structure optimization.
\end{IEEEbiography}

\begin{IEEEbiography}[{\includegraphics[width=1in,height=1.25in,clip,keepaspectratio]{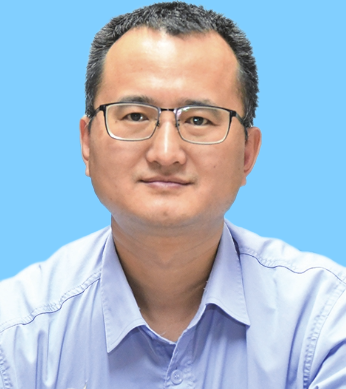}}]{Xinyu Bai} received his bachelor's degree and the master's degree from Liaoning Technical University, Fuxin, China, in 2004 and 2007, respectively. 
He is currently a principal engineer of AVIC Shenyang Aircraft Limited, where he has long been engaged in the development of process equipment and aircraft assembly. 
\end{IEEEbiography}

\vfill

\end{document}